\documentclass[lettersize,journal]{IEEEtran}
\usepackage{amsmath,amsfonts}
\usepackage{algorithmic}
\usepackage{algorithm}
\usepackage{array}
\usepackage[caption=false,font=normalsize,labelfont=sf,textfont=sf]{subfig}
\usepackage{booktabs}
\usepackage{textcomp}
\usepackage{stfloats}
\usepackage{url}
\usepackage{verbatim}
\usepackage{graphicx}
\usepackage{cite}
\usepackage{amssymb}
\usepackage{multirow}
\usepackage{booktabs}
\usepackage{graphicx}
\usepackage{color,xcolor}
\usepackage{hyperref}
\usepackage{threeparttable}
\usepackage{multirow}
\usepackage{amsmath}
\makeatletter
\renewcommand{\maketag@@@}[1]{\hbox{\m@th\normalsize\normalfont#1}}%
\makeatother

\begin{document}
\title{RSHazeDiff: A Unified Fourier-aware Diffusion Model for Remote Sensing Image Dehazing}

\author{Jiamei Xiong, Xuefeng Yan, Yongzhen Wang, Wei Zhao, Xiao-Ping Zhang, \textit{Fellow}, \textit{IEEE}, and Mingqiang Wei, \textit{Senior Member}, \textit{IEEE}

\thanks{Jiamei Xiong, Wei Zhao and Mingqiang Wei are with the School of Computer Science and Technology, Nanjing University of Aeronautics and Astronautics, Nanjing 210016, China (e-mail: jmxiong@nuaa.edu.cn, weizhao0120@nuaa.edu.cn, mingqiang.wei@gmail.com).}
\thanks{Xuefeng Yan is with the School of Computer Science and Technology, Nanjing University of Aeronautics and Astronautics, Nanjing 210016, and also with the Collaborative Innovation Center of Novel Software Technology and Industrialization, Nanjing 210093, China (e-mail: yxf@nuaa.edu.cn).}
\thanks{Yongzhen Wang is with the College of Computer Science and Technology, Anhui University of Technology, Ma’anshan 243099, China (e-mail: wangyz@ahut.edu.cn).}
\thanks{Xiao-Ping Zhang is with the Tsinghua Shenzhen International Graduate
School, Tsinghua University, Shenzhen, China. (e-mail: xpzhang@ieee.org).}
\thanks{\textit{(Corresponding author: Xuefeng Yan.)} }
}

\markboth{Journal of \LaTeX\ Class Files,~Vol.~14, No.~8, August~2021}%
{Shell \MakeLowercase{\textit{et al}}: A Sample Article Using IEEEtran.cls for IEEE Journals}


\maketitle

\begin{abstract}
Haze severely degrades the visual quality of remote sensing images and hampers the performance of road extraction, vehicle detection, and traffic flow monitoring. The emerging denoising diffusion probabilistic model (DDPM) exhibits the significant potential for dense haze removal with its strong generation ability. Since remote sensing images contain extensive small-scale texture structures, it is important to effectively restore image details from hazy images. However, current wisdom of DDPM fails to preserve image details and color fidelity well, limiting its dehazing capacity for remote sensing images. In this paper, we propose a novel unified Fourier-aware diffusion model for remote sensing image dehazing, termed RSHazeDiff. From a new perspective, RSHazeDiff explores the conditional DDPM to improve image quality in dense hazy scenarios, and it makes three key contributions. First, RSHazeDiff refines the training phase of diffusion process by performing noise estimation and reconstruction constraints in a coarse-to-fine fashion. Thus, it remedies the unpleasing results caused by the simple noise estimation constraint in DDPM. Second, by taking the frequency information as important prior knowledge during iterative sampling steps, RSHazeDiff can preserve more texture details and color fidelity in dehazed images. Third, we design a global compensated learning module to utilize the Fourier transform to capture the global dependency features of input images, which can effectively mitigate the effects of boundary artifacts when processing fixed-size patches. Experiments on both synthetic and real-world benchmarks validate the favorable performance of RSHazeDiff over state-of-the-art methods. Source code will be released at \textcolor{magenta}{\href{https://github.com/jm-xiong/RSHazeDiff}{https://github.com/jm-xiong/RSHazeDiff}}
\end{abstract}
\begin{IEEEkeywords}
RSHazeDiff, image dehazing, unified diffusion model, Fourier-aware refinement, phased training strategy           
\end{IEEEkeywords}

\section{Introduction}
\IEEEPARstart{R}{EMOTE} sensing (RS) records the surface information from high altitudes in a non-contact manner, providing a broader synoptic view for traffic monitoring with zero impact on traffic flow. Unfortunately, RS images captured by optical remote sensors inevitably suffer from noticeable degradation under hazy conditions. Such hazy images hamper visual tasks to extract traffic-related information from images with the help of intelligent transportation systems, such as road extraction \cite{zhang2017road, guan2021roadcapsfpn}, vehicle detection \cite{meng2023robust, hoanh2024multi}, traffic flow monitoring \cite{wang2019development, liu2022smart}, and autonomous driving \cite{han2021age}. RS image dehazing improves the overall visibility of degraded images and restores important traffic information, which reduces the risk of image misinterpretation. RS image dehazing is an essential prerequisite for reliable and precise performance of downstream vision-based transportation tasks.

As image dehazing is a typical ill-posed problem, early works exploit various prior assumptions to generate haze-free images \cite{he2010single, zhu2014single, xu2019iders, shen2020spatial, han2021edge}. The hand-crafted prior assumptions can regularize the solution space of linear inverse problems. However, predefined assumptions are only applied in specific scenarios, causing severe color shift and undesired artifacts in practice. Recently, numerous learning-based dehazing approaches have been proposed. Early efforts \cite{cai2016dehazenet, li2017aod, dong2020physics} develop convolutional neural networks (CNNs) to estimate the transmission maps or airlight, and generate sharp images via atmospheric scattering model (ASM) \cite{narasimhan2000chromatic}. However, the cumulative error and unreliable prior assumptions often cause color distortions or haze residuals. Since then, considerable end-to-end dehazing approaches \cite{liu2019griddehazenet, chen2021psd, cui2023focal, li2020coarse, zhang2022dense, bai2022self} directly generate haze-free images from the corresponding hazy images. With the advantage of generative models, some dehazing methods adopt generative adversarial networks (GANs) \cite{engin2018cycle, dong2020fd, pan2020physics, wang2022cycle, zheng2022dehaze} to regularize data distribution of dehazed images for haze removal. However, the training of GANs is unstable and susceptible to mode corruption.

Notably, natural image dehazing (NID) and remote sensing image dehazing (RSID) have several differences: First, RS images are taken from higher altitudes, so their surface information contains extensive small-scale texture details from land cover, vegetation, and urban buildings. Second, the atmospheric scattering model applied in NID does not work on RSID, due to the invariable scene depth in RS images \cite{zhang2022dense}. Third, RS images are usually captured from multiple bands. That is, RSID needs to consider the spectral characteristics of recovered images, especially when dealing with multispectral or hyperspectral data. Fourth, compared with the datasets for NID, there are limited datasets for RSID due to high available cost \cite{zhang2022dense}, thus it is valuable to construct large RS hazy datasets based on RS imaging mechanism.

\begin{figure*}[htbp] 
    \centering
    \includegraphics[width=1.0\linewidth]{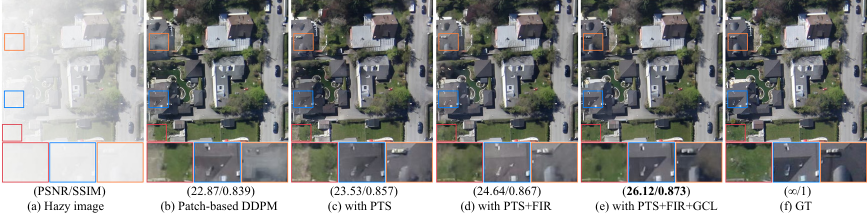}
	\caption{
	Impact of different components in RSHazeDiff. From (a) to (f): (a) the hazy image, and the dehazing results of (b) Patch-based DDPM model, (c) Patch-based DDPM model with PTS, (d) Patch-based DDPM model with PTS and FIR module, (e) our full model (Patch-based DDPM model + PTS + FIR + GCL), and (f) the ground-truth image. PTS, FIR and GCL refer to the phased training strategy, Fourier-aware iterative refinement module and global compensated learning module,  respectively.}
	\label{fig:fig1}
\end{figure*}

Diffusion model is an emerging generative paradigm and successful in various low-level vision tasks \cite{saharia2022image, saharia2022palette, whang2022deblurring}. With the superior modeling data distribution ability, diffusion models have achieved impressive performance in image generation. Moreover, the diffusion model is not subjected to unstable training and mode-collapse as GAN-based methods. Motivated by the strong generation ability of diffusion models, we leverage the conditional DDPM \cite{ho2020denoising} for RSID. However, there still exist several problems that limit the dehazing performance of DDPM. First, DDPM only uses simple noise-estimation constraint to optimize the network while ignoring image-level refinement. That simple constraining noise may introduce chaotic content and undesired impurity into restored images. Second, DDPM cannot utilize the potential structure and statistical information in the forward iterative process, limiting its better sampling capacity. During early iterations in the diffusion model, the forward iterative process preserves more structural and color information than reverse sampling process. It may provide additional beneficial information for finer sampling results, especially for dense haze removal. Note the early iterations mean that time step $t$ in DDPM is smaller. As shown in Fig. \ref{fig:fig1}(d), by exploiting the beneficial information in the forward iterative process, DDPM can better recover image details and color information. Third, although the patch-based DDPM enables size-agnostic processing and sampling acceleration, it yields evident block artifacts in that the patch-based strategy overlooks the global semantics of all patches. As shown in Fig. \ref{fig:fig1}(e)), introducing the global compensation learning module is beneficial to eliminate boundary artifacts and produce more visually pleasing images.

To address these issues, we propose a novel DDPM-based network for RSID, termed RSHazeDiff. Similar to the conditional DDPM, RSHazeDiff also adopts the degraded images as conditions to prevent content diversity with the stochastic sampling process. In addition, we develop a phased training strategy (PTS) to remedy the unsatisfactory results caused by simple noise-estimation constraint in the training phase. Concretely, we first train the diffusion model with the noise-estimation constraint, and then further optimize it with the reconstruction constraint. The reconstruction constraint aims to constrain the sampling results with its corresponding ground-truth images. Moreover, to take the most of latent structure and statistical information in the forward iterative process, we design a Fourier-aware iterative refinement (FIR) module. In the early sampling step (i.e., time step $t$ is small), FIR excavates the style/semantic information of the amplitude/phase components in Fourier domain for finer sampling results. Referring to the patch-based image processing in \cite{ozdenizci2023restoring}, we leverage the patch-based image processing strategy for unlimited-size image dehazing. However, separately dealing with each patch ignores the global dependency features of all patches, causing evident artifacts in the final result. In view of this, we utilize a global compensated learning module (GCL) to capture global dependency features. Comprehensive experiments verify the superiority of RSHazeDiff on both synthetic and real-world RS hazy images, especially in dense hazy scenes. 

Overall, our contributions are three-fold: 
\begin{itemize}
        \item A novel unified Fourier-aware conditional diffusion model is proposed for RSID (termed RSHazeDiff). It is a new exploration that applies diffusion models to RSID. RSHazeDiff leverages a phased training strategy to address the content inconsistency caused by the simple noise-estimation constraint, which is beneficial to produce more satisfactory dehazing results. 
        \item The Fourier-aware iterative refinement module is designed to further explore effective semantic and color information in the forward iterative process, better facilitating RSHazeDiff to preserve more texture details and color fidelity in restored images.   
        \item The global compensated learning module is introduced to eliminate evident block artifacts from the patch-based image processing scheme. By capturing the global dependency features in RS images, GCL assures the semantic consistency of all patches.
\end{itemize}

The remainder of this work is organized as follows. Section \ref{section II} presents the reviews of related work. Section \ref{section III} describes the preliminaries of the conditional DDPM. In Section \ref{section IV}, we introduce the general framework of RSHazeDiff and its details. Experimental results are presented and analyzed in Section \ref{section V}, followed by the conclusion in Section \ref{section VI}.

\section{Related Work}
\label{section II}
This section briefly reviews RSID algorithms and the diffusion model for image restoration, followed by the Fourier transform in deep learning.

\subsection{Remote Sensing Image Dehazing}
\textbf{Prior-based methods.} These methods \cite{xu2019iders, shen2020spatial, han2021edge} employ the hand-crafted priors to restore the dehazed images. For instance, Xu \textit{et al.} \cite{xu2019iders} design an iterative dehazing method named IDeRS, which introduces the concept of “virtual depth” for evaluating Earth surface coverings. Shen \textit{et al.} \cite{shen2020spatial} propose a spatial-spectral adaptive method for RSID, which corrects the ambient light and transmission estimation from the spatial and spectral dimensions. Han \textit{et al.} \cite{han2021edge} can effectively remove the haze of RS hazy images by exploiting multiscale guided filtering to decompose the input images into base layers and detail layers for final output. Although these prior-based methods achieve impressive results, they rely on prior assumptions in specific scenes, making the dehazing results easily violated when the assumptions are not satisfied.

\textbf{Learning-based methods.} With the success of deep learning, researchers have focused on learning-based RSID algorithms \cite{li2020coarse, zheng2022dehaze, zhang2022dense, liu2023multi, chi2023trinity, wang2023uscformer, liu2023nighthazeformer}. Similar to natural image
dehazing, some RSID methods explore deep neural networks to estimate haze parameters in ASM. For example, Chi \textit{et al.} \cite{chi2023trinity} develop a trinity model called Trinity-Net to incorporate prior knowledge into the gradient-guided Swin Transformer \cite{liu2021swin} for haze parameters estimation. The other works directly convert hazy images to haze-free images in an end-to-end manner. Among them, Li and Chen \cite{li2020coarse} design a coarse-to-fine dehazing network called FCTF-Net, in which the first stage exploits an encoder–decoder architecture to obtain coarse dehazing results and then refines them in second stage. Zhang and Wang \cite{zhang2022dense} propose a dynamic collaborative inference learning framework (DCIL) to handle haze removal of dense RS hazy images. Zheng \textit{et al.} \cite{zheng2022dehaze} exploit an enhanced attention-guide GAN-based network named Dehaze-AGGAN to restore haze-free remote sensing images in an unsupervised manner. However, few works focus on applying the diffusion model to image dehazing task, whereas the advanced generation ability of diffusion model makes it possible to better recover haze-free images. 

\subsection{Diffusion Models for Image Restoration}
Diffusion-based generative models \cite{sohl2015deep, ho2020denoising} progressively corrupt images by adding noise, and iteratively denoise from the noise distribution to sharp images. With the superior generative capability, diffusion-based models have achieved remarkable performance in image restoration, such as image super-resolution \cite{saharia2022image}, inpainting \cite{saharia2022palette}, and deblurring \cite{whang2022deblurring}. Saharia \textit{et al.} \cite{saharia2022image} adopt conditional diffusion model for image super-resolution and achieve more competitive results than GAN-based methods. With similar condition strategy, Saharia \textit{et al.} \cite{saharia2022palette} propose a unified framework to handle image-to-image translation tasks by incorporating distorted images as condition input. Rombach \textit{et al.} \cite{rombach2022high} apply diffusion models in latent space to improve sampling efficiency. Considering that current diffusion model architectures mainly focus on fixed-sized image restoration, Ozan \textit{et al.} \cite{ozdenizci2023restoring} propose a patch-based diffusion model for unlimited-size image restoration. However, it fails to capture global information when operating locally on fixed-sized patches, which yields apparent block artifacts. To address this issue, we design a global compensated learning module to capture long-range context in frequency domain.

\subsection{Fourier Transform in Deep Learning} 
Recently, the frequency information extracted through Fourier transform has attracted increasing attention in deep learning. A line of works utilizes it to improve representation ability of neural networks \cite{chi2020fast, wang2020high, zhou2022deep}. Among them, Wang \textit{et al.} \cite{wang2020high} reveal the relationship between the generalization of CNNs and the high-frequency components of images. Besides, Zhou \textit{et al.} \cite{zhou2022deep} propose deep Fourier up-sampling for multi-scale modeling, which achieves impressive performance on multiple computer vision tasks in a plug-and-play manner. The other line of work applies the Fourier transform to many computer vision tasks \cite{yang2020phase, yang2020fda, fuoli2021fourier, yu2022frequency, kong2023efficient, li2023embedding}. Taking image restoration tasks as an example, Fuoli \textit{et al.} \cite{fuoli2021fourier} adopt Fourier domain losses to better recover the high-frequency content in image super-resolution. Yu \textit{et al.} \cite{yu2022frequency} explore the haze degradation property from a frequency domain perspective to remove haze. Motivated by the convolution theorem that reveals an equivalent property between spatial domain and frequency domain, Kong \textit{et al.} \cite{kong2023efficient} propose an effective frequency domain-based Transformer for image deblurring, with less computational complexity. Inspired by some observations of low-light noisy images in the frequency domain, Li \textit{et al.} \cite{li2023embedding} design a cascaded network with Fourier Transform for low-light image enhancement. However, no works attempt to exploit the potential of the Fourier spectral components in diffusion models. To our knowledge, we are the first to take the Fourier spectral components as important prior knowledge during iterative sampling steps of diffusion model, better preserving image details and colors.

\section{Preliminaries}
\label{section III}
DDPM originates the diffusion model \cite{sohl2015deep}, which defines two essential processes, i.e., the forward/diffusion process and reverse/denoising process. For the conditional DDPM \cite{saharia2022image}, the degraded image is incorporated into the diffusion model as condition. With the guidance of conditional input, the diffusion model can pull the data distribution of restored images to conditional content distribution, preventing the content disparity caused by stochastic sampled noise. Next, we concisely introduce the conditional DDPM. 

\textbf{The diffusion process.} The forward process gradually coverts a clear data $x_{0}$ into a Gaussian noise distribution by $T$ iterations. For any iteration, the corresponding transformation is depicted as: 
\begin{equation}
\label{equ:equ1}
    q\left(x_{t}\mid x_{t-1}\right ) = \mathcal{N}\left (x_{t};\sqrt{1- \beta_{t}}x_{t-1}, \beta_{t}I\right) 
\end{equation}
where $\beta_{t}$ is the predefined variance schedule at the time step $t$, $x_{t}$ represents the noisy image, and $\mathcal{N}$ defines the Gaussian distribution. According to the property of the Markov chain, the above equation can be written in closed form:
\begin{align}
    q\left(x_{t}\mid x_{0}\right ) & = \mathcal{N}\left (x_{t};\sqrt{\overline{a_{t}}}x_{0}, \left ( 1-\overline{a_{t}}   \right ) I\right) \\
    &= \sqrt{\overline{a_{t}}}x_{0} + \sqrt{1-\overline{a_{t}}}\epsilon_{t}  
\end{align}
where $\overline{a_{t}} = \prod_{i=1}^{t}\left ( 1 -\beta _{i} \right )$, and $\epsilon_{t} \sim \mathcal{N}\left ( 0, \mathbf{I} \right )$ with the same dimensionality as $x_{t}$. When total step $T$ is large enough, $x_{t}$ follows a standard Gaussian distribution.  

\textbf{The denoising process.} The reverse process is designed to reconstruct the desired data samples from the Gaussian noise by progressively denoising based on Markov chain. Each denoising step of the reverse process can be described as: 
\begin{equation}
\label{equ:equ4}
    p_{0}\left ( x_{t-1}\mid x_{t}, \bar{x} \right ) = \mathcal{N}\left (x_{t-1}; u_{0}\left ( x_{t}, \bar{x}, t \right ), \sigma_{t}^{2} I\right )  
\end{equation}
where the invariant variance $\sigma_{t}^{2}= \frac{1-\bar{a}_{t-1} }{1-\bar{a}_{t}}\beta_{t}$, the learned mean $u_{0}\left ( x_{t}, \bar{x}, t \right )= \frac{1}{\sqrt{a_{t}}} \left ( x_{t}-\frac{1-a_{t}}{\sqrt{1-\bar{a}_{t}} }\epsilon_{0}\left ( x_{t}, \bar{x}, t\right )\right) $, and $\bar{x}$ defines the conditional image. In Eq. \ref{equ:equ4}, the $\epsilon_{0}\left ( x_{t}, \bar{x}, t\right ) $ is only variable and usually predicted by a denosing U-net network \cite{ronneberger2015u}. To approximate this value, the learned model is optimized by following training objective: 
\begin{equation}
\label{equ:equ5}
    \mathcal{L}_{noise} = E_{x_{0}, \bar{x}, t, \epsilon_{t}} \left [ \left \|\epsilon_{t} - \epsilon_{0} \left (\sqrt{\overline{a_{t}}}x_{0} + \sqrt{1-\overline{a_{t}}}\epsilon_{t}, \bar{x}, t\right ) \right \| ^{2} \right ]
\end{equation}

\begin{figure*}[t] 
    \centering
    \includegraphics[width=0.90\linewidth]{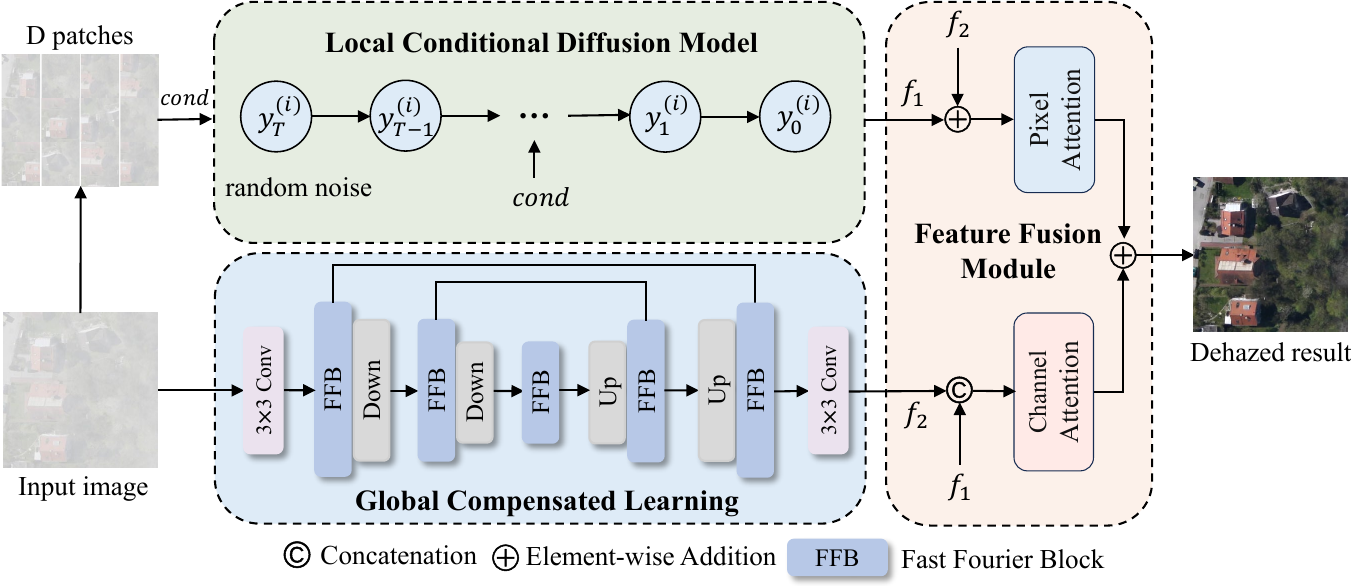}
	\caption{
	The overall architecture of RSHazeDiff for RSID. RSHazeDiff contains a local reverse denoising process, a global compensated learning module, and a feature fusion module. The local reverse denoising process focuses on locally sampling over patches, while GCL module performs the intact images with global properties in Fourier domain. Then, the global-local feature representations are fused by the feature fusion module to produce final dehazed images. We decompose an input image into $D$ overlapping fixed-sized patches via \cite{ozdenizci2023restoring}. $cond$ means that the $D$ patches are regarded as the conditions of local reverse denoising process. $y_{t}^{(i)}$ defines the $i$-th patch of the noisy image $y_{t}$ that gradually samples from the random noise patch $y_{T}^{(i)}$ to the haze-free patch $y_{0}^{(i)}$ by $t$ iterations. $f_{1}$ and $f_{2}$ represent the output of local reverse denoising process and GCL module, respectively.}
    \label{fig:fig2}
\end{figure*}

\section{Methodology}
\label{section IV}
We rethink RSID from the perspective of diffusion model. RSHazeDiff exploits the conditional DDPM to produce dehazing results with closer distribution to the clear data. Specifically, to eliminate that simple noise estimation produces unsatisfactory content in dehazing result, we leverage the reconstruction constraint to further optimize DDPM. Moreover, we explore the effective semantic and color information in the forward iterative process, better helping RSHazeDiff to preserve more texture details and color fidelity in sampled restoration results. Considering that patch-based image processing strategy cannot effectively capture global dependency features, we introduce a global compensated learning module to capture long-range context by executing simple yet effective spectrum transform in frequency domain.

In this section, we first describe the overview of RSHazeDiff. After that, the designed Fourier-aware iterative refinement and global compensated learning modules are elaborated in detail. Finally, we present the phased training strategy applied in the proposed method.

\subsection{Overview}
\begin{figure*}[htbp] 
    \centering
    \includegraphics[width=1.0\linewidth]{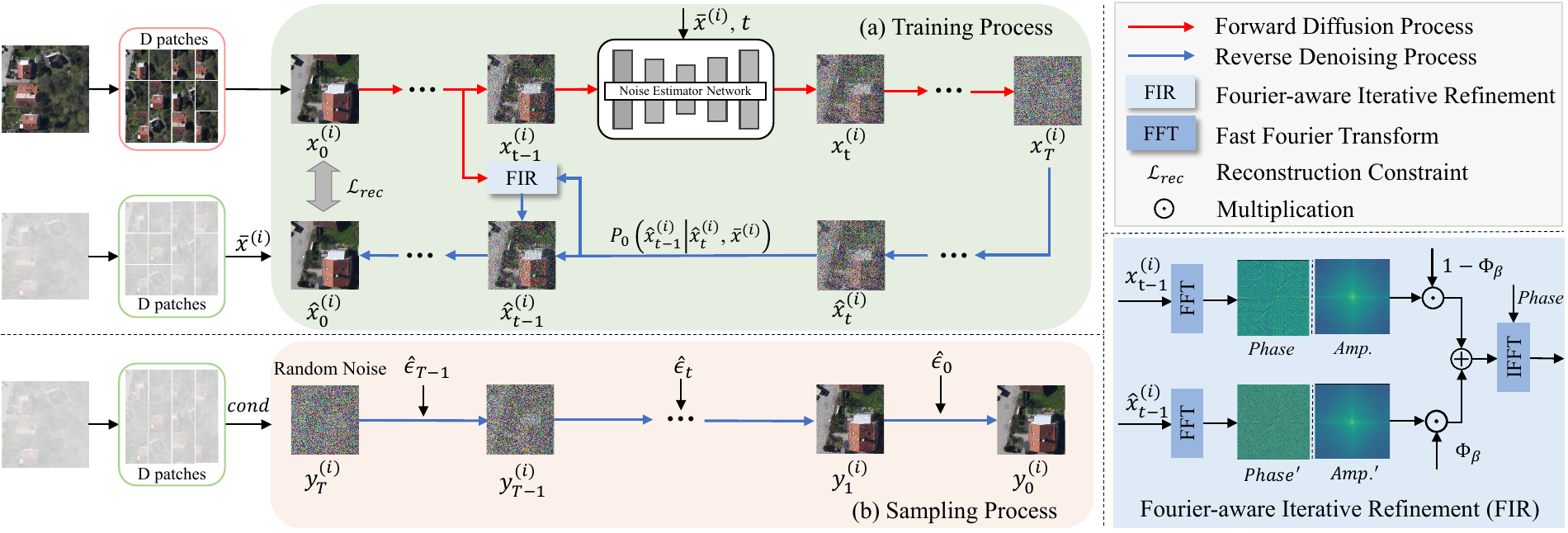}
	\caption{
	The architecture of the Fourier-aware conditional diffusion model, which contains two processes: (a) The training process adopts the phased training strategy to produce finer dehazing results. The model is first optimized with noise estimation constraint, and then the coarse sampling result by reverse denoising process is refined with reconstruction constraint. Also, the well-designed FIR excavates the structure and color guided representation from the forward process in Fourier space for finer sampling results. (b) The sampling process intends to gradually convert a Gaussian noise into the restored image. Note that the above model performs patch-level restoration. 
    }
    \label{fig:fig3}
\end{figure*}

\begin{figure*}[htbp] 
    \centering
    \includegraphics[width=0.75\linewidth]{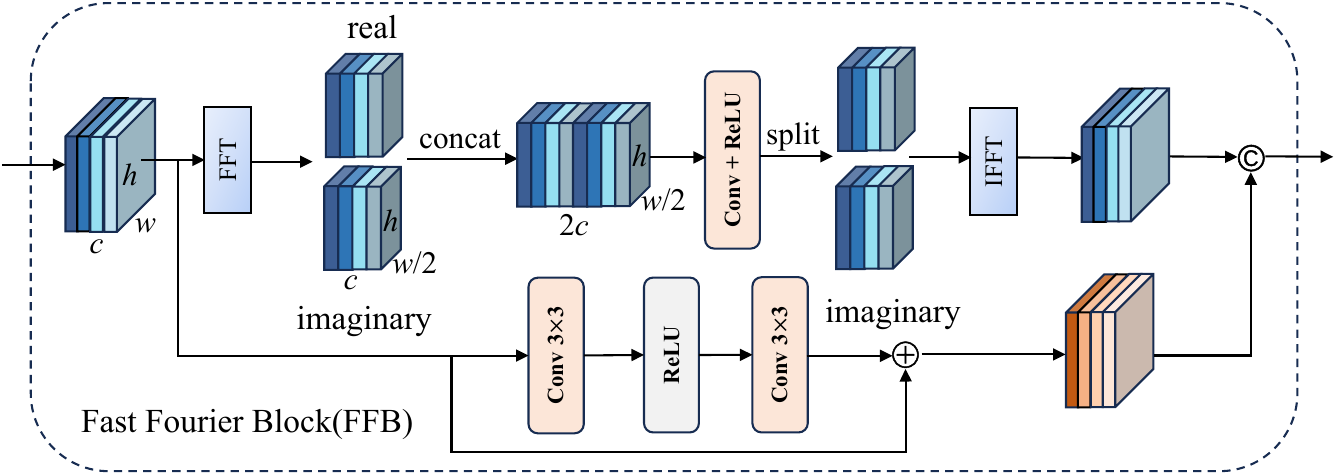}
	\caption{
	The detailed structure of the fast Fourier block, which can capture the long-range context by executing simple yet effective spectrum transform.
    }
    \label{fig:fig4}
\end{figure*}

Fig. \ref{fig:fig2} illustrates the overall architecture of RSHazeDiff, which consists of three main components: a local reverse denoising process, a global compensated learning module, and a feature fusion module. The local reverse denoising process leverages the Fourier-aware condition diffusion model to perform local patch-level restoration, and an input hazy image is randomly cropped with several patches as its conditions. The detailed training and sampling process of Fourier-aware conditional diffusion model is exhibited in Fig. \ref{fig:fig3}, and its pseudocodes are depicted in Alg. \ref{alg:training} and Alg. \ref{alg:inference}. Notably, during inference we crop the whole image into overlapping patches as the input of Fourier-aware conditional diffusion model by using the grid-like arranged parsing scheme \cite{ozdenizci2023restoring}. In turn, we merge all patches with the dictionary of their locations and adopt the mean estimated noise for overlapping pixels \cite{ozdenizci2023restoring} (lines 8,9,11 in Alg. \ref{alg:inference}). GCL module adopts the Fast Fourier Block (FFB) as the basic unit of U-net architecture to capture global feature representations. The structure of FFB is shown Fig. \ref{fig:fig4}. Finally, the feature fusion module fuses the global-local feature representations to produce the final dehazing results.

\begin{algorithm}[t]
    \caption{The training process of Fourier-aware conditional diffusion model}
    \label{alg:training}
    \textbf{Input}: Clean image $x_{0}$ , hazy image conditioner $\bar{x}$, denoiser network $\epsilon_{\theta}(x_{t},\bar{x},t)$, a random binary patch mask $\mathbf{P}_i$, transitional iteration $M$, number of implicit sampling steps $S$, time steps $T$, and the mask $\Phi_{\beta}$ in FIR module. 
    \begin{algorithmic}[1] 
        \WHILE {\textit{Not converged}}
            \STATE $x_{0}^{(i)}= \text{Crop}(\mathbf{P}_i \circ x_{0})$ and $\bar{x}^{(i)}=\text{Crop}(\mathbf{P}_i\circ \bar{x})$
            \STATE $t\sim \text{Uniform}\{1,\ldots,T\}$
            \STATE $\mathbf{\epsilon}_t \sim \mathcal{N}(\mathbf{0}, \mathbf{I})$
            \IF{\textit{Current iterations $\le M$}} 
                \STATE Take gradient descent step on 
                \STATE \qquad $\bigtriangledown_{\theta } \left \|\epsilon_{t} - \epsilon_{0} \left (\sqrt{\overline{a_{t}}}x_{0}^{(i)} + \sqrt{1-\overline{a_{t}}}\epsilon_{t}, \bar{x}^{(i)}, t\right ) \right \|^{2} $
            \ELSE
                \STATE $\hat{x}_{t_{S}}^{(i)} \sim {\mathcal{N} (0, \mathbf{I}})$
                \FOR {$j=S : 1$} 
                    \STATE $t = (j-1)\cdot T/S +1$
                    \STATE {$t_{next} = (j-2)\cdot
             T/S + 1$} \textbf{if} $j >1$ \textbf{else} $0$
                    \STATE $\hat{\epsilon}_{t} = \epsilon_{\theta}(\hat{x}_{t}^{(i)},\bar{x}^{(i)} ,t)$
                    \STATE $x_{t_{next}}^{(i)}= \sqrt{\overline{a}_{t_{next}}}x_{0}^{(i)} + \sqrt{1-\overline{a}_{t_{next}}}\hat{\epsilon}_{t_{next}}$
                    \STATE $\hat{x}_{t_{next}}^{(i)} =  \sqrt{\bar{\alpha}_{t_{next}}}(\frac{\hat{x}_{t}^{(i)}-\sqrt{1-\bar{\alpha}_{t}} \cdot \hat{\epsilon}_{t}}{\sqrt{\bar{\alpha}_{t}}})+\sqrt{1-\bar{\alpha}_{t_{next}}}\cdot \hat{\epsilon}_{t}$ 
                    \STATE $\hat{x}_{t_{next}}^{(i)} = FIR\left ( x_{t_{next}}^{(i)}, \hat{x}_{t_{next}}^{(i)}, \Phi_{\beta}\right )$
                    \ENDFOR
                \STATE Take gradient descent step on $\bigtriangledown_{\theta} \mathcal{L}_{rec} \left(\hat{x}_{0}^{(i)}, {x}_{0}^{(i)} \right )$
                \ENDIF
        \ENDWHILE
        \RETURN $\theta$
    \end{algorithmic}
\end{algorithm}
      
\begin{algorithm}[t]
    \caption{The sampling process of Fourier-aware conditional diffusion model}
    \label{alg:inference}
    \textbf{Input}: Hazy image $\bar{x}$, $\epsilon_{\theta}(y_{t},\bar{x},t)$, $S$, $T$, and dictionary of $D$ overlapping patch locations.
    \begin{algorithmic}[1] 
        \STATE $y_{t_{S}}\sim{\mathcal{N}(0,\mathbf{I}})$
        \FOR {$j=S : 1$} 
            \STATE $t = (j-1)\cdot T/(S-1) +1$
            \STATE {$t_{next}=(j-2)\cdot T/(S-1) + 1$} \textbf{if} $j >1$ \textbf{else} $0$
            \STATE $\mathbf{\hat{\Omega}}_t=\mathbf{0}$ and $\mathbf{M}=\mathbf{0}$
            \FOR {$d = 1 : D$}
            \STATE $ y_t^{(d)}=\text{Crop}(\mathbf{P}_d \circ y_t)$ and $\bar{x}^{(d)}=\text{Crop}(\mathbf{P}_d \circ \bar{x})$
            \STATE $\mathbf{\hat{\Omega}}_t = \mathbf{\hat{\Omega}}_t + \mathbf{P}_d \cdot \mathbf{\epsilon}_{\theta}(y_t^{(d)}, \bar{x}^{(d)},t)$
            \STATE $\mathbf{M} = \mathbf{M} + \mathbf{P}_d$
            \ENDFOR
            \STATE $\mathbf{\hat{\Omega}}_t = \mathbf{\hat{\Omega}}_t \oslash \mathbf{M}$\qquad\quad$\mathbin{/\mkern-4mu/}$\;\,$\oslash$: element-wise division
            \STATE $y_{next}=\sqrt{\bar{\alpha}_{t_{\text{next}}}}\left(\frac{y_t-\sqrt{1-\bar{\alpha}_t}\, \cdot\, \mathbf{\hat{\Omega}}_t}{\sqrt{\bar{\alpha}_t}}\right) + \sqrt{1-\bar{\alpha}_{t_{\text{next}}}}\cdot \mathbf{\hat{\Omega}}_t$
        \ENDFOR
        \STATE \textbf{return} $y_{0}$
    \end{algorithmic}
\end{algorithm}

\subsection{Fourier-aware Iterative Refinement}
DDPM-based image restoration approaches tend to tackle the forward diffusion process and reverse denoising process in sequence. The reverse process progressively sample with variable $\epsilon_{0}\left ( x_{t}, \bar{x}, t\right)$ predicted in the forward process. Apart from that, the two processes are independent to a certain extent. Nevertheless, early iterations in the forward process can provide beneficial guided features to the sampling process. As depicted in \cite{xu2021fourier}, the amplitude and phase components in Fourier space correspond to low-level statistics and high-level semantic information of input images. This characteristic can be extended in image dehazing task. The amplitude spectrum contains more hazy degradation properties (e.g., illumination and contrast), while the phase spectrum preserves more texture structure of images. In early sampling steps (time step $t$ is small), dense hazy images suffer from severe structure and color information loss, whereas the forward iterative process preserves more structural and color information. Inspired by that, we design a Fourier-aware iterative refinement module, which excavates the structure and color guided representation from the forward process in Fourier domain to refine transitions of sampling process. The well-designed FIR module can help diffusion model to generate more natural and realistic images.   

The detailed structure of FIR is illustrated in Fig. \ref{fig:fig3}. Note that the $x_{0}^{(i)}$ represents $i$-th patch cropped from the clear image $x_{0}$. we transition step from a patch $x_{t-1}^{(i)}$ to $x_{t}^{(i)}$ through Eq. \ref{equ:equ1}, and then sample from $\hat{x}_{t}^{(i)}$ to $\hat{x}_{t-1}^{(i)}$ through Eq. \ref{equ:equ4}. Then, images $x_{t-1}^{(i)}$ and $\hat{x}_{t-1}^{(i)}$ are decomposed to amplitude and phase components in Fourier domain by Fast Fourier Transform (FFT) \cite{prince1994fast}. Next, we use the phase spectrum of $x_{t-1}^{(i)}$ and integrate the partial amplitude spectrum of $x_{t-1}^{(i)}$ into amplitude component of $\hat{x}_{t-1}^{(i)}$. Finally, we apply the inverse Fast Fourier Transform (IFFT) to transform the final fused results $\hat{x}_{t-1}^{(i)}$ back to spatial domain image. Given a single channel image $x\in\mathcal{R}^{H \times W}$, its Fourier transform $\mathcal{F}$ as a complex component can be expressed as:
\begin{equation}
    \mathcal{F}\left ( x \right ) \left ( m,n \right ) = \frac{1}{\sqrt{HW} } \sum_{h=0}^{H-1}\sum_{w=0}^{W-1}x\left ( h,w \right )e^{-j^{2 \pi} \left ( \frac{h}{H}m+ \frac{w}{W}n \right ) }          
\end{equation}
here $m,n$ define the coordinates in Fourier space, and $j^{2} = -1 $. For an RGB image, each channel is adopted the Fourier transform separately. The amplitude component $\mathcal{A}\left ( x \right )$ and phase component $\mathcal{P}\left ( x \right )$ can be computed as:
\begin{equation}
    \mathcal{A}\left ( x \right ) \left ( m, n \right ) = \sqrt{R ^{2} \left ( x \right )\left ( m, n \right ) + I ^{2} \left ( x \right )\left ( m, n \right )} 
\end{equation}
\begin{equation}
    \mathcal{P}\left ( x \right ) \left ( m, n \right ) = \arctan \left [ \frac{I \left ( x \right )\left ( m, n \right )}{R \left ( x \right )\left ( m, n \right )}  \right ]
\end{equation}
where $R\left ( x \right )$ and $I\left ( x \right )$ are the real and imaginary parts of the complex component. Therefore, every reverse step can be formalized as follows:
\begin{equation}
    \hat{x}_{t-1}^{\left ( i \right ) }  = \mathcal{F}^{-1} \left ( \Phi_{\beta }  \mathcal{A}\left ( \hat{x}_{t-1}^{\left ( i \right ) } \right ) + \left ( 1- \Phi_{\beta } \right ) \mathcal{A}\left ( x_{t-1}^{\left ( i \right ) } \right ), \mathcal{P} \left ( x_{t-1}^{\left ( i \right ) } \right )   \right )  
\end{equation}
where $\hat{x}_{t-1}^{\left(i\right)}=\text{Crop}\left ( \mathbf{P}_i \circ \hat{x}_{t-1} \right )$ and $x_{t-1}^{\left(i\right)}=\text{Crop}\left ( \mathbf{P}_i \circ x_{t-1} \right )$ define $i$-th $p\times p$ patch from image pair $\left ( \hat{x}_{t-1}, x_{t-1} \right ) $. The $\text{Crop}\left ( \circ  \right ) $ operation extracts the patch from the location $\mathbf{P}_i$ within the complete image. $\mathcal{F}^{-1}$ defines inverse Fourier transform. Similar to \cite{yang2020fda}, $\Phi_{\beta}$ represents an image mask that only the value of a small center region is non-zero. The small center area is $\left [ -\beta H: \beta H , -\beta W: \beta W \right ] $. The coordinates of the center in the images are set $\left ( 0, 0 \right )$ and $\beta \in \left ( 0, 1 \right )$.

By fully exploiting the unexplored texture structure and color information in forward iterative process, FIR module can refine DDPM to generate more visually pleasing images. The relevant experiments in Section \ref{section:Ablation} show that dehazing performance is improved with the use of the FIR.

\subsection{Global Compensated Learning}
It is significant that the image restoration methods can handle unlimited-size images in practice. To achieve unlimited-size image restoration, we adopt the patch-based processing strategy that divides images into multiple fixed-size patches and then deals with each patch independently. However, this patch-based strategy may produce artifacts with the ignorance of global dependency information in all patches. To address this, we propose an effective GCL module that adopts FFB as the basic U-net architecture unit to capture all patches' global dependency information. After learning global feature representations, the designed fusion module fuses the local patch-based image restoration process and GCL module in a pixel-channel attention collaborative manner. The structure of GCL module and the fusion module are illustrated in Fig. \ref{fig:fig2}. 

In GCL, we apply FFB to deal with the entire images based on the spectral convolution theorem \cite{katznelson2004introduction}, which observes that updating each pixel value in spectral space affects all spatial pixels (i.e., Fourier domain embraces global properties). As illustrated in Fig. \ref{fig:fig4}, FFB employs the $1\times 1$ convolution operator to tackle real and imaginary parts in Fourier space for capturing frequency representations, while adopts a residual block with two 3 × 3 convolution layers for capturing spatial representations. After that, the final results are produced by concatenation and convolution operation. Therefore, the simple yet effective spectrum transform in the Fourier space can be used to capture the long-range context feature.

After capturing the global feature, we use a feature fusion module to fuse local patch-based denoising process and global compensated learning module. Referring to the statement in \cite{dong2020multi}, only adopting addition or concatenation operations to fuse two feature maps is less effective. Hence, we introduce two parallel attention modules to further fuse features in a pixel-channel collaborative manner. The pixel attention (PA) \cite{qin2020ffa} can produce a $1 \times H \times W$ attention map while the channel attention (CA) \cite{qin2020ffa} produces a $C\times 1\times 1$ attention map. The two attention modules can handle different types of information flexibly, i.e., haze pixels and more important channel information. Therefore, we adopt the element-wise addition operation between the local patch-based denoising process and GCL module before PA and the channel-wise concatenation operation before CA. Above all, the collaborative pixel-channel fusion is used to generate the final dehazing result. The fusion module can be described as follows:
\begin{equation}
    \mathcal{F}_{out} = \mathcal{C} \left ( CA\left ( \left [\mathcal{C}\left ( f_{1}\right ),\mathcal{C}\left(f_{2}\right) \right ] \right ) + PA\left (\mathcal{C}\left(f_{1}\right) + \mathcal{C}\left(f_{2}\right) \right )  \right )  
\end{equation}
where $f_{1}$ and $f_{2}$ are the output of local reverse denoising process and global compensated learning module, respectively. $\mathcal{C}$ represents convolution layer and $\left [\cdot\right]$ is concatenation operation.

\subsection{Phased Training Strategy}
Simple constraining the training phase of diffusion model with noise estimation cannot effectively restore satisfactory results. That is because the noise-estimation constraint focuses on distribution-closer approximation, whereas narrowing the disparity between restored and clear data distribution cannot enable model to generate visually pleasing output. In view of this, we employ reconstruction constraint to refine the training phase to obtain finer result with realistic colors and rich details. Specifically, PTS optimizes DDPM by constraining the noise-estimation in the early stages. After that, a coarse restored image, obtained by performing the sampling algorithm in the training phase, is constrained with corresponding ground-truth ones for finer restored results. PTS refines unsatisfactory results caused by simple noise-estimation constraint.    

The noise-estimation constraint is depicted in Eq. \ref{equ:equ5}. The $L_1$ loss and multi-scale structural similarity loss (MS-SSIM) are adopted as the reconstruction constraint. Since \cite{zhao2016loss} has proved that $L_1$ loss can provide better convergence and restoration performance than $L_2$ loss, we leverage $L_1$ loss to constrain the coarse restored results with ground-truth ones. The MS-SSIM loss ($\mathcal{L}_{ms-ssim}$) focuses on assessing the difference between the coarse restored results $\hat{x}_{0}^{(i)}$ and clear patch $x_{0}^{(i)}$ and its can be defined as follows: 
\begin{equation}
    \begin{split}
    & \mathcal{L}_{ms-ssim}= 1 - \\
    & \prod_{m=1}^{M}\left ( \frac{2u_{\hat{x}_{0}^{(i)}}u_{x_{0}^{(i)}}+C_{1}}{u_{\hat{x}_{0}^{(i)}}^{2}+u_{x_{0}^{(i)}}^{2}+C_{1}} \right )^{\beta _{m}}\left ( \frac{2\sigma _{\hat{x}_{0}^{(i)}\cdot x_{0}^{(i)}}+C_{2}}{\sigma _{\hat{x}_{0}^{(i)}}^{2}+\sigma _{x_{0}^{(i)}}^{2}+C_{2}} \right )^{\gamma_{m}}
\end{split}
\end{equation}
here $u_{\hat{x}_{0}^{(i)}}$, $u_{x_{0}^{(i)}}$ and $\sigma _{\hat{x}_{0}^{(i)}}$, $\sigma _{x_{0}^{(i)}}$ define the mean and standard deviations of $\hat{x}_{0}^{(i)}$ and $x_{0}^{(i)}$, and $\sigma_{\hat{x}_{0}^{(i)}\cdot x_{0}^{(i)}}$ is their covariance. $C_{1}$ and $C_{2}$ are constants and usually set $C_{1}=0.0001$, $C_{2}=0.0009$, $\beta_{m}$ and $\gamma_{m}$ are the relative importance of two components and both set to 1. $M$ represents the number of scales and we set it to five according to \cite{wang2003multiscale}. Therefore, the overall reconstruction constraint $\mathcal{L}_{rec}$ during the late training stage is defined as: 
\begin{equation}
    \mathcal{L}_{rec} = \mathcal{L}_{1} + \mathcal{L}_{ms-ssim}
\end{equation}

In general, the conditional DDPM is trained by simple noise-estimation constraint ($\mathcal{L}_{noise}$) in the early stages, and then minimizes the reconstruction constraint ($\mathcal{L}_{1}$ and $\mathcal{L}_{ms-ssim}$) for finer training optimization. Note that these loss functions are locally operated on fixed-sized patches.

\textit{Remark:} The noise-estimation constraint is necessary for the local conditional diffusion model. In the reverse process, the noise subtracted at each time step is estimated by the U-Net network. The network can predict the noise $\epsilon_{0}\left ( x_{t}, \bar{x}, t\right )$ with the noise-estimation constraint (Eq. \ref{equ:equ5}), further obtaining the mean and variance of $x_{t-1}$ (Eq. \ref{equ:equ4}). If we remove the noise-estimation constraint in the early stages (i.e., only the reconstruction constraints used in the training stages of local conditional diffusion model), the noise prediction network is invalid, causing severe degradation of sampling quality. In this case, only using the reconstruction constraints makes it difficult to predict the dehazed images from input images with Gaussian noise distribution. When we only use the reconstruction constraints in the whole training stages, the dehazed results are pure noisy images even training for 30w iterations in the ERICE dataset. It further validates our analysis.

Apart from the above mentioned loss function that is optimized over local patch, we also exploit the perceptual loss to global compensated learning module for constraining the whole image. The perceptual loss function can measure high-level perceptual similarity between the output and ground truth by extracting the multi-layer features from pre-trained network VGG-19 \cite{simonyan2014very}. By minimizing perceptual loss, the restored output $f_{2}$ has more similar texture and structure features with ground-truth images. Similar to \cite{johnson2016perceptual}, we hierarchically extract  five latent feature maps from VGG-19 model to measure the image perceptual similarity between the output $f_{2}$ and ground-truth images $z$, which can be expressed as:
\begin{equation}
\mathcal{L}_{perceptual}=\sum_{i=1}^{5}\omega_{i}\left \| F_{i}\left ( f_{2} \right ) - F_{i}\left ( z  \right )  \right \| _{1}    
\end{equation}
where $F_{i}\left( \cdot\right )$ represents the $i$-th latent features obtained by VGG-19. $\omega_{i}$ denotes the weight coefficient of the $i$-th latent features. Here, we obtain latent features from layers 1, 6, 11, 20, and 29 in the VGG-19 model, and the $\omega_{i}$ ($i$ from 1 to 5) are set to $\frac{1}{32}, \frac{1}{16}, \frac{1}{8}, \frac{1}{4}, 1$, respectively. Furthermore, we also employ pixel-level $\mathcal{L}_{1}$ loss to assess the diversity between $f_{2}$ and $z$. Hence, the loss function $\mathcal{L}_{global}$  in global compensated learning branch is shown as: 
\begin{equation}
    \mathcal{L}_{global} = \mathcal{L}_{1} + \mathcal{L}_{perceptual}
\end{equation}

Overall, PTS enables diffusion model to effectively fine local details in patch-based manner. In addition, we employ perceptual loss to optimize the final result by narrowing the semantic disparity between restored results and ground-truth image. In this way, RSHazeDiff can learn global-local feature representations to produce visually more pleasing results.

\section{Experiments}
\label{section V}
In this section, the implementation details of our experiments are first described. Next, the comparisons with several recent approaches on the synthetic and real-world datasets are performed. After that, a spectral consistency analysis is conducted to verify the spectral similarity between restored images and reference images. Then, we conduct ablation studies to demonstrate the effectiveness of core components in RSHazeDiff, followed by the application of our method in intelligent transportation systems. Finally, we discuss the limitation of our method and future work.

\subsection{Implementation Details}
\textbf{Dataset}. We evaluate the performance of RSHazeDiff on three RS hazy datasets, \textit{i.e.}, the light hazy image dataset (LHID), the dense hazy image dataset (DHID) \cite{zhang2022dense} and expanded remote sensing image cloud removing dataset (ERICE) \cite{zhang2022dense}. LHID and DHID are synthetic datasets, and each image has 512$\times$512 pixels. LHID contains 30,517 image pairs for training and 1000 for testing (LHID-A and LHID-B). The DHID dataset contains 14,490 image pairs for training and 500 for testing. The DHID dataset is generated by randomly sampling the ambient light $A\in \left [ 0.7, 1 \right ]$ and 500 different real-world transmission maps. The ERICE is a real-world dataset expanded on RICE1 \cite{lin2019remote}, collected in some scene categories from Google Earth, such as mountain, river, and desert. In ERICE dataset, 1,850 and 150 images are regarded as training and test sets, each with a resolution of 256$\times$256 pixels. Additionally, we employ the NHM dataset \cite{zhang2020nighttime} to evaluate the nighttime dehazing capability of RSHazeDiff. Like \cite{zhang2020nighttime}, the training set is 8,073 image pairs from NHR dataset\cite{zhang2020nighttime}.

\begin{table*}[htbp]
	\centering
	\caption{Quantitative comparisons with 17 dehazing approaches on the synthetic LHID and DHID datasets. The $\uparrow$ means higher is better while $\downarrow$ means lower is better. The \textcolor{red}{red} and \textcolor{blue}{blue} colors are used to indicate the $1^{st}$ and $2^{nd}$ ranks, respectively}
	\label{Tab:1}
	\begin{threeparttable}
		\footnotesize
		\centering
		\setlength{\tabcolsep}{0.8mm}{
			\begin{tabular}{cccccccccccc}
				\toprule
				\multirow{2}{*}{Method}&
				\multirow{2}{*}{Publication}&
				\multicolumn{5}{c}{LHID}&
				\multicolumn{5}{c}{DHID}\cr
				 \cmidrule(lr){3-7} \cmidrule(lr){8-12}&
				 & PSNR$\uparrow$ & SSIM$\uparrow$ & CIEDE2000$\downarrow$ & LPIPS$\downarrow$ & FID$\downarrow$ & PSNR$\uparrow$ & SSIM$\uparrow$ & CIEDE2000$\downarrow$  & LPIPS$\downarrow$ & FID$\downarrow$ \cr
				\midrule
				DCP \cite{he2010single}   & TPAMI'10 & 21.12 & 0.818 & 9.161 & 0.481 & 21.49 & 18.92 & 0.824 & 9.867 & 0.163 & 79.92  \cr
				AOD-Net \cite{li2017aod}   & ICCV'17  & 22.24 & 0.842 & 7.122 & 0.475 & 32.61 & 15.37 & 0.718 & 14.272 & 0.269 & 151.16 \cr
                GridDehazeNet \cite{liu2019griddehazenet}   & ICCV'19 & 26.64 & 0.883 & 4.597 & 0.451 & 18.86 & 25.51 & 0.868 & 5.053 & 0.121 & 65.95 \cr
                IDeRs \cite{xu2019iders} & INS'19  & 14.68 & 0.627 & 19.563 & 0.633 & 44.65 &  13.33 & 0.575 & 19.942 & 0.344 & 120.17 \cr
                FFA-Net \cite{qin2020ffa} & AAAI'20  & 27.79 & 0.887 & 4.424 & 0.451 & 16.27 & 25.05 & 0.880 & 5.590 & 0.144 & 73.27 \cr
			PhysicsGAN \cite{pan2020physics} &  
                TPAMI'20  & 22.01 & 0.835 & 8.198 & 0.512 & 38.72 & 19.85 & 0.824 & 10.415 & 0.187 & 94.61\cr
			PSD \cite{chen2021psd} & CVPR'21 &           23.33 & 0.855 & 6.998
                & 0.549 & 37.25 & 23.30 & 0.853 & 6.911 & 0.163 & 82.66 \cr
                FCTF-Net \cite{li2020coarse}  & GRSL'21 & 25.85 & 0.876 & 5.423 & 0.459 & 21.21 & 17.21 & 0.739 & 12.794 & 0.221 & 128.49 \cr
			DCIL \cite{zhang2022dense}  & TGRS'22 & 28.12 & 0.898 & 5.180  & 0.447 & 14.93 & 26.95 & \textcolor{blue}{0.896} & 4.877 & 0.131 & 64.73 \cr
                SGID-PFF \cite{bai2022self} &       
            TIP'22  & 24.12 & 0.852 & 7.091 & 0.488 & 30.13 & 25.69 & 0.867 & 6.147 & 0.158 & 80.74 \cr
			Trinity-Net \cite{chi2023trinity} & TGRS'23  & 24.78 & 0.867 & 6.737 & 0.460 & 17.76 & 25.86 & 0.878 & 5.479 & 0.139 & 75.46 \cr
                RIDCP \cite{wu2023ridcp} &       
            CVPR'23  & 26.75 &  0.843 & 4.553 & \textcolor{blue}{0.128} & 19.07 & 26.15 & 0.874 & 5.329 & 0.136 & 64.36 \cr
                C$^{2}$PNet \cite{zheng2023curricular} &   
            CVPR'23  & 28.65 & 0.867  & 4.065 & \textcolor{red}{0.086} & 15.16 & 26.20 & 0.871 & 4.925 & \textcolor{blue}{0.106} & 56.40 \cr
                Focal-Net \cite{cui2023focal} & ICCV'23  & \textcolor{blue}{29.39} & 0.899 & \textcolor{blue}{4.012} & 0.445 & \textcolor{blue}{14.32} & 25.73 & 0.887 & 5.313 & 0.131 & 73.04 \cr
                PSMB-Net \cite{sun2023partial} & 
            TGRS'23  & 28.89 & \textcolor{blue}{0.900} & 5.864 & 0.454 & 16.13 & \textcolor{blue}{27.07} & 0.887 & 4.965 & 0.113 & \textcolor{blue}{53.83} \cr
                AU-Net \cite{du2024dehazing} &    
            TGRS'24  & 28.44 & 0.893 & 4.045 & 0.451 & 16.96 & 26.42 & 0.891 & \textcolor{blue}{4.489} & 0.138 & 69.28 \cr
			Ours & - & \textcolor{red}{29.65} &              \textcolor{red}{0.905} & \textcolor{red}{3.875} & 0.436 & \textcolor{red}{12.85} &        
                \textcolor{red}{27.91} & \textcolor{red}{0.900} & \textcolor{red}{3.945} & \textcolor{red}{0.102} & \textcolor{red}{49.52} \cr
				\bottomrule
			\end{tabular}
		}
	\end{threeparttable}
\end{table*}

\textbf{Training Details}. All experiments are implemented by PyTorch \cite{paszke2019pytorch} framework with NVIDIA GeForce RTX 3090 GPU. In the training of Fourier-aware conditional diffusion model, we adopt the Adam ($\beta_{1}=0.9$, $\beta_{2}=0.999$) optimizer with a batch size of 4. The initial learning rate is set to 0.00002. The single input image is randomly cropped to 16 patches for training, each patch with a resolution of 64$\times$64. We adopt PTS to train the model until it converges. Concretely, we first train the model with noise estimation constraint for 1,500,000 iterations, and then refine it with reconstruction constraint for 500,000 iterations in LHID. Similarly, we first train the model for 1,000,000 iterations and then refine it for 500,000 iterations in DHID. To expedite the sampling process, we adopt the implicit sampling strategy \cite{song2020denoising} in which the total number of time steps and implicit sampling steps are set to 1,000 and 10. Furthermore, we use the exponential moving average (EMA) strategy \cite{song2020improved} to eliminate slight color shift problem. During inference, the single input image is decomposed into $D$ overlapping patches, each with a resolution of 64$\times$64. $D$ denotes the number of overlapping patches and is set to 64. For the training of the global compensated learning module, we use the Adam optimizer with the batch size of 8 and the initial learning rate of 0.001. 

\textbf{Evaluation Settings}. To demonstrate the superiority of RSHazeDiff, we qualitatively and quantitatively compare it against 16 representative dehazing algorithms, including DCP \cite{he2010single}, AOD-Net \cite{li2017aod}, GridDehazeNet \cite{liu2019griddehazenet}, IDeRs \cite{xu2019iders}, FFA-Net \cite{qin2020ffa}, PhysicsGAN \cite{pan2020physics}, PSD \cite{chen2021psd}, FCTF-Net \cite{li2020coarse}, DCIL \cite{zhang2022dense}, SGID-PFF \cite{bai2022self}, Trinity-Net \cite{chi2023trinity}, RIDCP \cite{wu2023ridcp}, C$^{2}$PNet \cite{zheng2023curricular}, Focal-Net \cite{cui2023focal}, PSMB-Net \cite{sun2023partial}, and AU-Net \cite{du2024dehazing}. 

We adopt PSNR \cite{assessment2004error}, SSIM \cite{assessment2004error}, and CIEDE2000 \cite{sharma2005ciede2000} as the distortion metrics and LPIPS \cite{zhang2018unreasonable} and FID \cite{heusel2017gans} as the perceptual metrics for dehazing performance evaluation. PSNR index assesses the error between the dehazed images and the corresponding ground-truth ones pixel by pixel. SSIM measures image similarity regarding brightness, contrast, and structure. CIEDE2000 indicator is employed to evaluate the color distinction between the dehazed image and its haze-free counterpart. LPIPS and FID assess the perceptual quality and realism of the dehazed images. Larger values of PSNR and SSIM indicate better results, while smaller values of CIEDE2000, LPIPS, and FID indicate better performance.

\begin{figure*}[htbp] \centering
 	\includegraphics[width=1.0\linewidth]{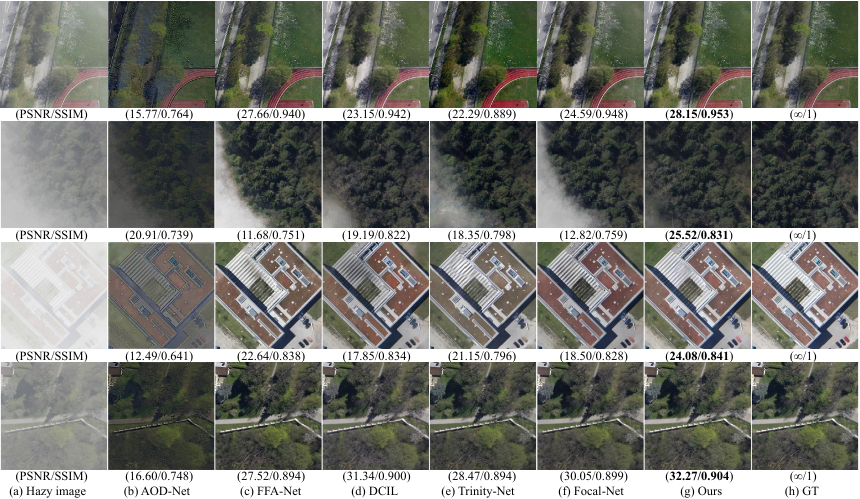}
 	\caption{Qualitative comparisons on DHID dataset. From (a) to (h): (a) the hazy image, and the dehazing results of (b) AOD-Net\cite{li2017aod}, (c) FFA-Net \cite{qin2020ffa}, (d) DCIL \cite{zhang2022dense}, (e) Trinity-Net \cite{chi2023trinity}, (f) Focal-Net \cite{cui2023focal}, (g) our RSHazeDiff, respectively, and (h) the ground-truth image. As observed, our RSHazeDiff can generate much
    clearer haze-free images with well-preserved details.}
   \label{fig:fig5}
\end{figure*}

\begin{figure*}[htbp] 
  \centering
	\includegraphics[width=1.0\linewidth]{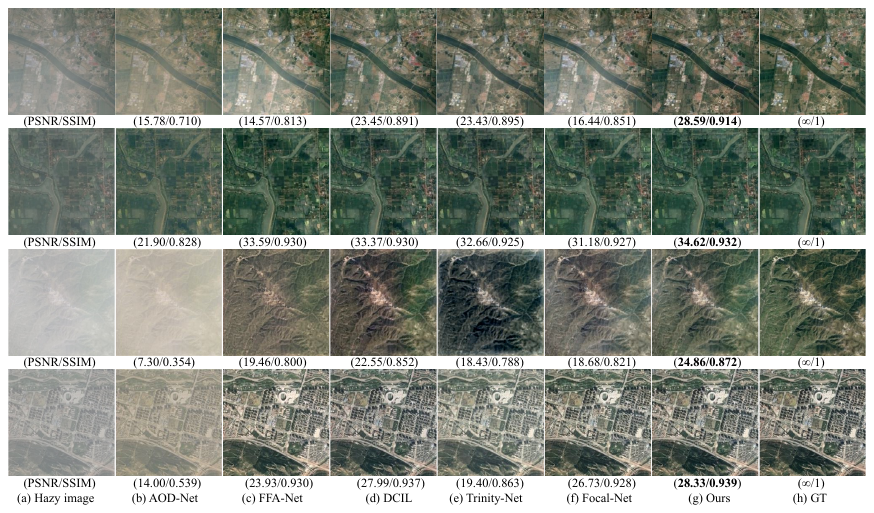}
	\caption{Qualitative comparisons on the real-world ERICE dataset. From (a) to (h): (a) the hazy image, and the dehazing results of (b) AOD-Net\cite{li2017aod}, (c) FFA-Net \cite{qin2020ffa}, (d) DCIL \cite{zhang2022dense}, (e) Trinity-Net \cite{chi2023trinity}, (f) Focal-Net \cite{cui2023focal}, (g) our RSHazeDiff, respectively, and (h) the ground-truth image. Our RSHazeDiff can generate more natural images with realistic colors and rich details.}
	\label{fig:fig6}
\end{figure*}

\subsection{Comparison with State-of-the-art Methods}
\textbf {Results on Synthetic Dataset.} Table \ref{Tab:1} reports the quantitative performance of 17 dehazing algorithms on synthetic LHID and DHID test datasets. To make fair comparisons, we re-train all compared methods on synthetic datasets according to the implementation details of their papers. Note that we directly re-train the whole network of DCIL on DHID dataset without the phased learning strategy \cite{zhang2022dense}. As observed, our RSHazeDiff can achieve the highest PSNR and SSIM values and the lowest CIEDE2000 and FID values than other representative dehazing algorithms, which indicates that our RSHazeDiff is able to better restore the texture details and color information from hazy RS images. Moreover, Table \ref{Tab:2} offers comparisons with several nighttime dehazing methods on the NHM dataset. As observed, RSHazeDiff wins first and second place among ten representative nighttime dehazing methods, revealing its potential in nighttime dehazing tasks.

\begin{table}[htbp]
	\centering
	\caption{Quantitative comparisons on the nighttime hazy dataset NHM}
	\label{Tab:2}
	\begin{threeparttable}
		\centering
		\setlength{\tabcolsep}{0.8mm}{
			\begin{tabular}{cccc}
				\toprule
				\multirow{1}{*}{Method}&
				\multirow{1}{*}{Publication}&
                \multirow{1}{*}{PSNR$\uparrow$}&
                \multirow{1}{*}{SSIM$\uparrow$} \cr
				\midrule
                    NDIM \cite{zhang2014nighttime} & ICIP'14 & 12.69 & 0.617 \cr
                    GS \cite{li2015nighttime} & ICCV'15 & 11.85 & 0.615 \cr
                    MRP \cite{zhang2017fast} & CVPR'17 & 13.11 & 0.655 \cr
                    FAST-MRP \cite{zhang2017fast} & CVPR'17 & 13.31 & 0.649 \cr
                    OSFD \cite{zhang2020nighttime} & MM'20 & 13.28 & 0.667 \cr 
                    VD \cite{liu2022nighttime} & CVPRW’22 & 13.77 & 0.687 \cr 
                    GHLP \cite{wang2022variational} & TIP’22 & 13.15 & 0.674 \cr 
                    GAPSF \cite{jin2023enhancing} &       
                    MM’23 & 13.48 & 0.665 \cr
                    NightHazeFormer \cite{liu2023nighthazeformer} & MM’23 & \textcolor{red}{18.54} & \textcolor{blue}{0.784} \cr
				Ours & - & \textcolor{blue}        {15.86} & \textcolor{red}{0.786} \cr
				\bottomrule
			\end{tabular}
		}
	\end{threeparttable}
\end{table}

Fig. \ref{fig:fig5} presents the visual comparisons against several recent dehazing approaches on DHID datasets. We can observe that AOD-Net tends to darken the whole images and produce noticeable Halo artifacts, degrading the overall visibility of the dehazed images. FFA-Net and Focal-Net cannot remove haze effectively when dealing with dense hazy images, leading to some haze residuals. In addition, the dehazing results of FFA-Net suffer from slight color distortion. Although DCIL can remove a majority of haze to a certain extent, it may distort some texture details of input images and fails to completely remove the haze in some regions. Trinity-Net cannot produce pleasing dehazing results due to haze residuals and serious color distortion. Compared with above dehazing algorithms, our RSHazeDiff can generate much clearer haze-free images with realistic color fidelity and finer texture details.

\textbf{Results on Real-World Hazy Images.} To further validate the dehazing performance on real-world images, we evaluate our method against several recent dehazing approaches on ERICE dataset. The ERICE dataset contains some images captured in thick haze scenarios. Table \ref{Tab:3} exhibits the quantitative comparison results of different dehazing algorithms. As reported, the proposed method surpasses other dehazing algorithms in terms of PSNR and SSIM, indicating that the images restored by RSHazeDiff are most similar to its corresponding ground-truth images. Although CIEDE2000 value of RSHazeDiff is not the best, our method wins the second place among eleven competitive dehazing algorithms. The quantitative results verify that the superiority of RSHazeDiff on real-world RS images.

\textit{Remark:} The above experiment applies the model trained on the training set of ERICE dataset to evaluate the performance of real-world haze removal. To further evaluate the generalization ability of our model,  we train the model on the synthetic dataset DHID and test it on an unseen real-world dataset ERICE. Compared to Table \ref{Tab:3}, all methods suffer from obvious performance degradation, typically because of the domain gap between synthetic and real-world hazy images. Comfortingly, our method wins first place in these indicators. To avoid repetition, we just exhibit the results in Table \ref{Tab:3}.

\begin{table}[htbp]
	\centering
	\caption{Average PSNR, SSIM and CIEDE2000 of various different approaches on the real-world ERICE dataset}
	\label{Tab:3}
	\begin{threeparttable}
		\centering
		\setlength{\tabcolsep}{0.8mm}{
			\begin{tabular}{ccccc}
				\toprule
				\multirow{1}{*}{Method}&
				\multirow{1}{*}{Publication}&
                \multirow{1}{*}{PSNR$\uparrow$}&
                \multirow{1}{*}{SSIM$\uparrow$}&
                \multirow{1}{*}{CIEDE2000$\downarrow$} \cr
				\midrule
				DCP \cite{he2010single}  & TPAMI'10 & 18.13 & 0.816 & 11.482 \cr
				AOD-Net \cite{li2017aod}  & ICCV'17 & 20.18 & 0.807 & 9.438 \cr
                GridDehazeNet \cite{liu2019griddehazenet} &  ICCV'19 & 32.73 & 0.935 & 2.719 \cr
                IDeRs \cite{xu2019iders} & INS'19 & 13.29 & 0.513 & 20.701 \cr
				FFA-Net \cite{qin2020ffa}  & AAAI'20 & 30.81 & 0.932 & 3.475 \cr
				PSD \cite{chen2021psd}  & CVPR'21 & 26.58 & 0.918 & 5.218 \cr
                FCTF-Net \cite{li2020coarse}  & GRSL'21 & 33.13 & 0.933 & 3.626 \cr
				DCIL \cite{zhang2022dense}  & TGRS'22 & \textcolor{blue}{35.90} & \textcolor{blue}{0.942} & \textcolor{red}{2.068} \cr
				Trinity-Net \cite{chi2023trinity}  & TGRS'23  &  30.26 & \textcolor{blue}{0.942} & 4.026 \cr
                Focal-Net \cite{cui2023focal}  & ICCV'23  &  32.67 & 0.937 & 2.691 \cr
				Ours & - & \textcolor{red}{36.56} & \textcolor{red}{0.958} & \textcolor{blue}{2.642} \cr
				\bottomrule
			\end{tabular}
		}
	\end{threeparttable}
\end{table}

Fig. \ref{fig:fig6} presents four real-world hazy samples from ERICE and the corresponding dehazing results produced by different dehazing methods. As displayed, AOD-Net fails to produce visually pleasing images due to abundant haze residuals, especially in thick hazy scenarios. The dehazed images of FFA-Net and Focal-Net still remain residual haze and cannot restore clear texture details. Although the dehazing results produced by DCIL seem good, they suffer from slight color shift and contain residual haze in some regions. Trinity-Net fails to remove haze completely and may yield over-saturation in restored images. In contrast, our method succeeds in removing the haze of input images completely. Furthermore, our RSHazeDiff can generate visually more pleasing images with faithful colors and richer structural information.

\begin{figure}[htbp] \centering
 	\includegraphics[width=1.0\linewidth]{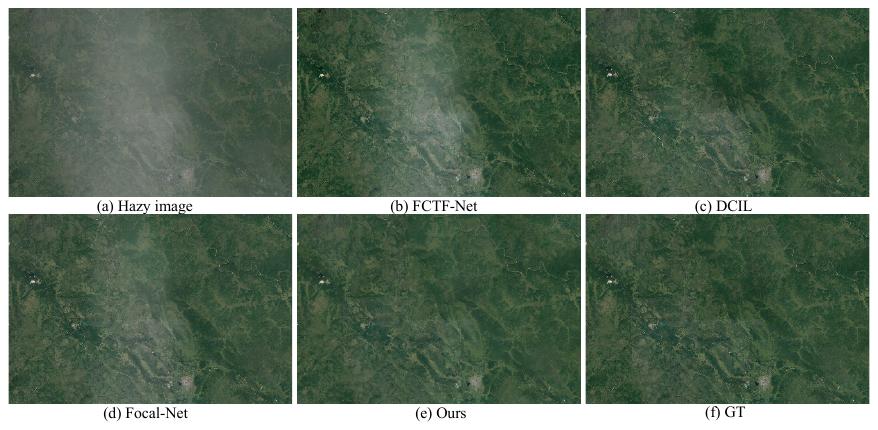}
 	\caption{Dehazing results of different methods on the full-size RICE1 dataset. From (a) to (f): (a) the hazy image with nonuniform haze (3072 × 2048), and the dehazing results of (b) FCTF-Net \cite{li2020coarse}, (c) DCIL \cite{zhang2022dense}, (d) Focal-Net \cite{cui2023focal}, (e) our RSHazeDiff, respectively, and (f) the ground-truth image.}
   \label{fig:fig7}
\end{figure}

Fig. \ref{fig:fig7} exhibits the comparison results of different dehazing methods on the full-size RICE1 dataset, with the size of 3072 × 2048. As a display, FCTF-Net has poor ability in handling nonuniform hazy images, leading to abundant haze residual. Focal-Net still cannot remove most of nonuniform haze. The dehazing result of DCIL contains slight haze residual in some regions. In contrast, our RSHazeDiff can remove the nonuniform haze completely.

\begin{figure}[htbp] \centering
 	\includegraphics[width=1.0\linewidth]{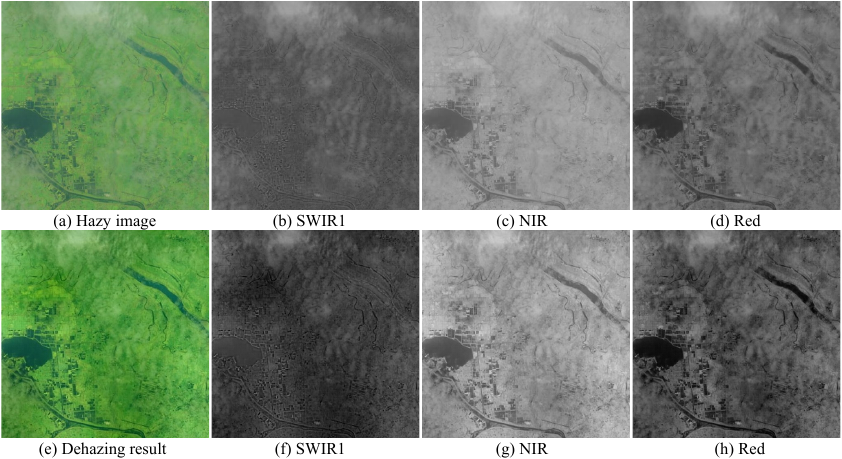}
 	\caption{Dehazing result of each band on RICE2 dataset \cite{lin2019remote}. From (a) to (d): (a) the hazy image composed by three bands, (b) the shortwave infrared 1 (1.57-1.65 $\mu$m), (c) the near-infrared band (0.85-0.88 $\mu$m), and (d) the red band (0.64-0.67 $\mu$m). (e)-(h) are the dehazing results of (a)-(d), respectively.}
   \label{fig:fig8}
\end{figure}

We also evaluate the dehazing performance of our method on other bands. The hazy samples on RICE2 dataset \cite{lin2019remote} are composed of three bands, \textit{i.e.}, shortwave infrared 1 (SWIR 1) band, near-infrared (NIR) band, and red band. Fig. \ref{fig:fig8} shows the dehazing results of our method among the three bands. As observed, RSHazeDiff can produce clear images with rich structures and few haze residuals.

\textbf{Generalization to out-of-distribution degradations.} To show the generalization ability of RSHazeDiff, we train our model on the synthetic dataset DHID and test it on the training-unseen datasets D-Hazy \cite{ancuti2016d} and T-Cloud \cite{ding2022uncertainty}. The two datasets have different haze distribution and climatic conditions from the training set. The quantitative evaluation results are shown in Table \ref{Tab:4}. As reported, we win first place in terms of PSNR, SSIM, and CIEDE2000, and also achieve impressive performance in terms of LPIPS and FID, which further verifies the favorable generalization ability of our method.

\begin{table*}[htbp]
	\centering
	\caption{Quantitative comparisons with recent dehazing approaches on the D-Hazy and T-Cloud test set}
	\label{Tab:4}
	\begin{threeparttable}
		\footnotesize
		\centering
		\setlength{\tabcolsep}{1.0mm}{
			\begin{tabular}{ccccccccccc}
				\toprule
				\multirow{2}{*}{Method}&
				\multicolumn{5}{c}{D-Hazy}&
				\multicolumn{5}{c}{T-Cloud}\cr
				 \cmidrule(lr){2-6} \cmidrule(lr){7-11}&
			PSNR$\uparrow$ & SSIM$\uparrow$ & CIEDE2000$\downarrow$ & LPIPS$\downarrow$ & FID$\downarrow$ & PSNR$\uparrow$ & SSIM$\uparrow$ & CIEDE2000$\downarrow$  & LPIPS$\downarrow$ & FID$\downarrow$ \cr
				\midrule
			AOD-Net \cite{li2017aod}  & \textcolor{blue}{13.55} &      0.648 & \textcolor{blue}{15.060} & \textcolor{blue}{0.262} & 41.39 &        19.32 & 0.602 & 11.974 & \textcolor{red}{0.317} &        \textcolor{red}{111.38} \cr			
                FFA-Net \cite{qin2020ffa}  & 11.48	& 0.648 & 19.817 & 0.314 & \textcolor{blue}{34.20} & 19.27 & 0.619 & 12.637 & 0.392& 138.79 \cr 
			DCIL \cite{zhang2022dense} & 11.03	      & 0.686 & 19.801 & 0.304 &              \textcolor{red}{33.23}       &          \textcolor{blue}{20.08} &               \textcolor{blue}{0.716} &               \textcolor{blue}{10.590} & 0.369        & 145.43 \cr
                RIDCP \cite{wu2023ridcp} & 12.20 & 0.677	& 18.778	& 0.311	& 34.70 & 19.71	& 0.661	& 10.989	& 0.338	& \textcolor{blue}{114.39} \cr
                Focal-Net \cite{cui2023focal}  & 
                8.92	 & 0.608	 & 26.116 & 	0.409	 & 40.59 &  18.57	 & 0.655	 & 11.364 & 	0.352	 & 116.60 \cr
                PSMB-Net \cite{sun2023partial} & 
                13.14	& 0.636	& 18.641	& 0.447	& 79.28 & 20.05	& 0.709	& 16.861& 	0.457	& 122.37 \cr
                AU-Net \cite{du2024dehazing} & 
                13.12	& \textcolor{blue}{0.717}	& 15.351	& 0.289	& 37.30 & 19.52	& 0.692 &	11.903	& 0.400	& 135.40 \cr
			Ours  & 
                \textcolor{red}{13.66}	& \textcolor{red}{0.725}	& \textcolor{red}{15.008}	& \textcolor{red}{0.236}	& 39.75 & \textcolor{red}{20.35}	& \textcolor{red}{0.739}	& \textcolor{red}{10.444} & 	\textcolor{blue}{0.337}	& 114.54 \cr
				\bottomrule
			\end{tabular}
		}
	\end{threeparttable}
\end{table*}

\subsection{Analysis of Spectral Consistency}
We conduct a spectral consistency analysis to further evaluate the spectral similarity between recovered image and its corresponding ground-truth image in typical land covers. The spectral angle mapper (SAM) \cite{yuhas1992discrimination} index can calculate the spectral similarity between dehazed images $\bar{y}$ and reference images $y$, which is formulated as:
\begin{equation}
SAM = \frac{1}{P} \sum_{p=1}^{P}\cos^{-1} \left ( \frac{y_{p}^{T} \bar{y}_{p}}{\left \| y_{p} \right \| _{2}\cdot \left \| \bar{y}_{p} \right \| _{2}  }   \right ) 
\end{equation}
where $P$ is the total number of pixels. $y_{p}$ and $\bar{y}_{p}$ are the vectors $\in \mathcal R^{1\times C }$ for the pixel $p$. The $C$ means the number of bands. A smaller value of SAM indicates higher spectral similarity between dehazed images and reference images.

\begin{figure}[htbp] \centering
 	\includegraphics[width=0.9\linewidth]{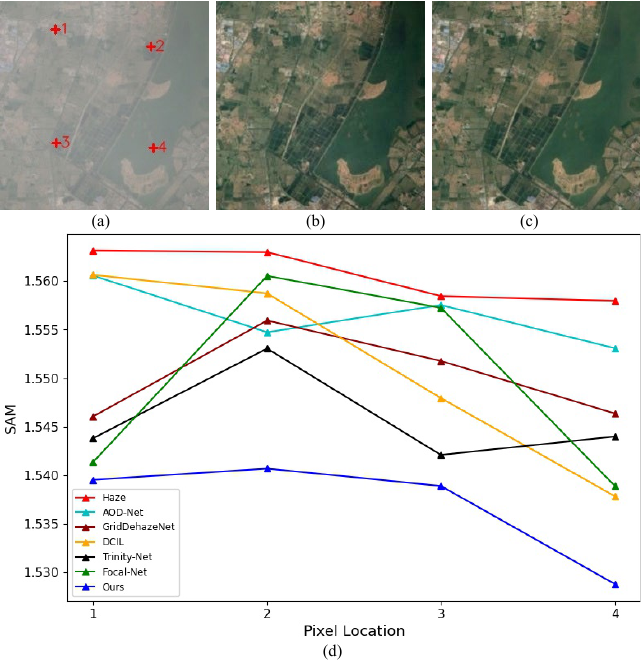}
 	\caption{Spectral profiles of different land covers on ERICE dataset. (a) the hazy image, (b) the dehazing result of our RSHazeDiff, (c) the ground-truth image, and (d) spectral comparison results for the pixels masked by cross 1-4 in the hazy image.}
   \label{fig:fig9}
\end{figure}

As exhibited in Fig. \ref{fig:fig9}, we report SAM values of different dehazing methods in some land covers marked by cross 1-4 (e.g., terrace, stream, and soil). As observed, SAM values of haze images among the four land covers are the largest. Compared with other algorithms, our RSHazeDiff achieves the smallest SAM values, which indicates our recovered spectral features of typical land covers are most similar to the ground-truth image.

\subsection{Ablation Study}
\label{section:Ablation}
\textbf{Effect of different components in RSHazeDiff.} Our method shows dehazing superiority both quantitatively and qualitatively over 17 representative dehazing methods. Furthermore, we perform an ablation study to analyze the effectiveness of the proposed components, including the phased training strategy, Fourier-aware iterative refinement, and global compensated learning module. 

We first perform the baseline model with patch-based conditional DDPM and then optimize this model with noise-estimation constraint. Next, key modules are progressively added into the baseline model:
\begin{enumerate}
\item  baseline + phased training strategy $\rightarrow$ $V_1$, 
\item  $V_1$ + Fourier-aware iterative refinement  $\rightarrow$ $V_2$,
\item  $V_2$ + global compensated learning module $\rightarrow$ $V_3$ (our full model).
\end{enumerate}
For fair comparisons, all variants are trained in the same way as before and tested on DHID dataset. The performances of these variants can be found in Table \ref{Tab:5} and Fig. \ref{fig:fig1}.

\begin{table}[htbp]
	\centering
	\footnotesize
	\caption{Ablation study of different components on DHID dataset.}
	\label{Tab:5}
	\begin{threeparttable}
		\centering
		\setlength{\tabcolsep}{1.3mm}{
			\begin{tabular}{ccccc}
				\toprule
		    	\multirow{1}{*}{Variants}&
				\multirow{1}{*}{Baseline}&
				\multirow{1}{*}{$V_1$}&
				\multirow{1}{*}{$V_2$}&
				\multirow{1}{*}{$V_3$}\cr
				\midrule
				Phased training strategy & w/o & \checkmark & \checkmark & \checkmark  \cr
				Fourier-aware iterative refinement & w/o & w/o  & \checkmark & \checkmark  \cr
				Global compensated learning module & w/o & w/o & w/o & \checkmark \cr
				\midrule
				PSNR & 25.50 & 26.09 & 26.86 &  \textbf{27.91} \cr
				SSIM & 0.878 & 0.881 & 0.892 & \textbf{0.900} \cr
                CIEDE2000 & 6.176 & 5.044 & 4.352 & \textbf{3.945} \cr
				\bottomrule
			\end{tabular}
		}
	\end{threeparttable}
\end{table}

As shown in Table \ref{Tab:5} and Fig. \ref{fig:fig1}, each component of RSHazeDiff can boost the dehazing performance. When applying the phased training strategy, the model achieves advanced performance that has fewer haze residuals and higher metric values over baseline model. Furthermore, the introduction of FIR and GCL module also have brought performance improvement of model in three  indicators. In short, the full model performs favorably against the other variants both quantitatively and qualitatively, demonstrating that all the components of RSHazeDiff are beneficial and necessary for effective haze removal.

\textbf{Effect of sampling steps in sampling process.} During the sampling process, diffusion model iteratively denoises until generating sharp results through $S$ steps. Theoretically, a large sampling step contributes to producing more visually pleasing result \cite{whang2022deblurring}, whereas it also increases the inference time. To further explore the performance of different sampling steps, we perform an ablation study setting different $S$ from 5 to 25. Table \ref{Tab:6} exhibits the average PSNR, SSIM and CIEDE2000 of different sampling steps. As reported, we can find that the performance degrades slightly as the number of sampling step is greater than 10. To carefully balance the performance and sampling time, we set $S$ to 10 in our experiment.

\begin{table}[htbp]
	\centering
	\footnotesize
	\caption{Ablation study on the sampling steps in sampling process (hyperparameter $S$)}
	\label{Tab:6}
	\begin{threeparttable}
		\centering
		\setlength{\tabcolsep}{1.2mm}{
			\begin{tabular}{ccccccccc}
				\toprule
				\multirow{1}{*}{}&
				\multirow{1}{*}{$S=5$}&
                \multirow{1}{*}{$S=8$}&
				\multirow{1}{*}{$S=10$}&
				\multirow{1}{*}{$S=12$}&
				\multirow{1}{*}{$S=15$}&
                \multirow{1}{*}{$S=25$}\cr
				\midrule
				PSNR & 26.38 & 27.12 & \textbf{27.91} & 26.97 & 26.85 & 27.64 \cr
				SSIM & 0.900 & \textbf{0.902} & 0.900 & 0.900 & 0.897  & 0.886\cr
				CIEDE2000 & 4.394 & 4.473 & \textbf{3.945} & 4.238 & 4.290 & 4.552\cr
				\bottomrule
			\end{tabular}
		}
	\end{threeparttable}
\end{table}

\textbf{Effect of the mask scale in FIR.} The mask scale (hyperparameter $\beta$) is the weight of different amplitude components in FIR. Obviously, when the hyperparameter $\beta$ is set to 0, the Fourier-aware iterative refinement module is invalid. When $\beta$ is set to 1, FIR focuses on fusing the phase component of forward process and amplitude component of reverse process in Fourier domain. Table \ref{Tab:7} shows the dehazing performance of different scale values from 0.01 to 1. As reported, FIR can better help diffusion model to restore haze-free images when the $\beta$ is set to 0.1.

\begin{table}[htbp]
	\centering
	\footnotesize
	\caption{Ablation study on the mask scale in FIR (hyperparameter $\beta$)}
	\label{Tab:7}
	\begin{threeparttable}
		\centering
		\setlength{\tabcolsep}{1.2mm}{
			\begin{tabular}{cccccccc}
				\toprule
				\multirow{1}{*}{}&
				\multirow{1}{*}{$\beta=0.01$}&
				\multirow{1}{*}{$\beta=0.05$}&
				\multirow{1}{*}{$\beta=0.1$}&
                \multirow{1}{*}{$\beta=0.5$}&
                \multirow{1}{*}{$\beta=1$}\cr
				\midrule
				PSNR & 27.11 & 27.33 & \textbf{27.91} & 26.97 & 27.67 \cr
				SSIM & 0.888 & 0.887 & \textbf{0.900} & 0.886 & 0.890 \cr
                CIEDE2000 & 4.963 & 4.958 & \textbf{3.945} & 5.188 & 4.047 \cr
				\bottomrule
			\end{tabular}
		}
	\end{threeparttable}
\end{table}

\textbf{Effectiveness of the local-global feature representations.} We compare our method with the representative global modeling-based image restoration approaches on the ERICE dataset, including TransWeather \cite{valanarasu2022transweather}, Fourmer \cite{zhou2023fourmer}, and FSNet \cite{cui2023image}. As exhibited in Table \ref{Tab:8}, our method is far superior to other methods in four evaluation metrics. In detail, our method surpasses the TransWeather, Fourmer, and FSNet methods by 2.64$\%$, 4.44$\%$, and 1.89$\%$ in PSNR, respectively.

\begin{table}[htbp]
	\centering
	\footnotesize
	\caption{Performance comparisons of different global modeling-based methods on the ERICE dataset}
	\label{Tab:8}
	\begin{threeparttable}
		\centering
		\setlength{\tabcolsep}{1.0mm}{
			\begin{tabular}{ccccccc}
				\toprule
				\multirow{1}{*}{}&
				\multirow{1}{*}{TransWeather \cite{valanarasu2022transweather}}&
				\multirow{1}{*}{Fourmer  \cite{zhou2023fourmer}}&
				\multirow{1}{*}{FSNet \cite{cui2023image}}&
                    \multirow{1}{*}{Ours}\cr
				\midrule
				PSNR & 33.92 & 32.12 & 34.67 & \textbf{36.56}\cr
				SSIM & 0.934 & 0.932 & 0.941 & \textbf{0.958} \cr
                    LPIPS & 0.053 & 0.074 & 0.059 & \textbf{0.052} \cr
                    FID & 41.66 & 59.29 & 36.94 & \textbf{32.97} \cr
				\bottomrule
			\end{tabular}
		}
	\end{threeparttable}
\end{table}

\subsection{Application}
RS image dehazing provides clear observation of transportation infrastructure, such as roads, vehicles, tunnels, bridges, and buildings. That enables downstream visual-based transportation tasks to extract useful traffic information from images under hazy conditions. The downstream transportation applications include road extraction, urban road planning, traffic congestion monitoring, vehicle detection, infrastructure development, and disaster response. Here we take vehicle detection as an example to exhibit the applicability of our approach in transportation field.

\begin{figure*}[htbp] 
  \centering
	\includegraphics[width=0.95\linewidth]{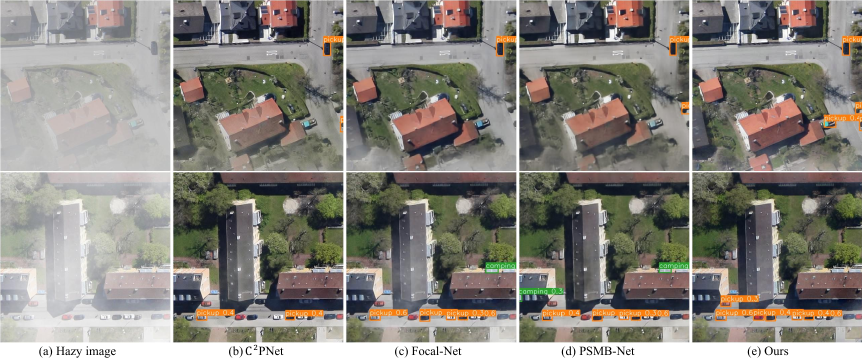}
	\caption{Vehicle detection results on the hazy images and  images after dehazing by different methods. From (a) to (e): (a) vehicle detection results
on the hazy images, and the vehicle detection results after dehazing by (b) C$^{2}$PNet \cite{zheng2023curricular}, (c) Focal-Net \cite{cui2023focal}, (d) PSMB-Net \cite{sun2023partial}, and (e) our RSHazeDiff, respectively.}
	\label{fig:fig10}
\end{figure*}

Vehicle detection helps road network planning and transport traffic statistics, widely used in vision-based intelligent transportation systems, yet its detection accuracy is noticeably degraded under hazy conditions. To verify that our RSHazeDiff can benefit vehicle detection, we employ a pre-trained YOLOv5s \cite{jocher2022ultralytics} detector to detect the vehicles on the original hazy images and the dehazing results by different methods. The YOLOv5s detector is retrained on the fine-grained vehicle detection dataset VEDAI \cite{razakarivony2016vehicle} to ensure stable vehicle detection. Fig. \ref{fig:fig10} exhibits two samples from the dense hazy images DHID. As shown, after dehazing, the detection performance is significantly improved. Compared to other dehazing algorithms, our method succeeds in removing thick haze with sharper structures and details, and achieves superior accuracy for vehicle detection.

\subsection{Limitation and Future Work}
Although RSHazeDiff achieved impressive performance, our method requires longer inference time compared to end-to-end image dehazing networks, due to the multiple iterative evaluations and gradient calculations in DDPM. Even if we adopt the DDIM-based sampling strategy to reduce the number of sampling steps to 10, the inference speed of our method is still slower than the end-to-end dehazing approaches. Our RSHazeDiff takes 1.016 seconds to restore an image of size 256 × 256 on a single NVIDIA GeForce RTX 3090 GPU. Recently, the latent diffusion model \cite{rombach2022high} performs the diffusion process in a compressed latent space rather than the pixel space. It will be a potential direction to improve the computational efficiency of our method. In future work, we will explore the dehazing performance of diffusion model in the lower dimension latent space.

\section{Conclusion}
\label{section VI}
In this paper, we propose an effective unified Fourier-aware diffusion model for RSID, namely RSHazeDiff. Our work offers a new perspective to tackle image dehazing task, which exploits the generative ability of the conditional DDPM to improve image perceptual quality in dense hazy scenarios. Different from the primitive DDPM, RSHazeDiff performs a fine training phase to pull the sampled dehazed images closer to their corresponding ground-truth counterparts, which can evade the unsatisfactory results caused by simple noise estimation. In addition, we design a Fourier-aware iterative refinement module to help the diffusion model to preserve more realistic colors and richer texture details in the restored image. Furthermore, to mitigate boundary artifacts when processing diffusive patches, a well-designed global compensated learning module is developed to capture the global dependency features of input images. Quantitative and qualitative experiments validate that RSHazeDiff is superior to SOTAs on dense RS image datasets.

\vspace{5pt}
\noindent\textbf{Acknowledgements.}
The authors thank the editors and anonymous reviewers for their careful reading and valuable comments. 

\bibliographystyle{IEEEtran}
\bibliography{reference}

\vfill
\end{document}